\documentclass{article}
\usepackage{ijcai24}

\usepackage{times}
\usepackage{soul}
\usepackage{url}
\usepackage[hidelinks]{hyperref}
\usepackage[utf8]{inputenc}
\usepackage[small]{caption}
\usepackage{graphicx}
\usepackage{amsmath}
\usepackage{amsthm}
\usepackage{booktabs}
\usepackage{algorithm}
\usepackage{algorithmic}
\usepackage[switch]{lineno}

\usepackage{multicol}
\usepackage{multirow}

\title{Empirical Analysis of Dialogue Relation Extraction with Large Language Models}

\author{
    Anonymous submission
}

\author{
Guozheng Li$^1$\and
Zijie Xu$^1$\and
Ziyu Shang$^1$\and
Jiajun Liu$^1$\and
Ke Ji$^1$\And
Yikai Guo$^2$\\
\affiliations
$^1$School of Computer Science and Engineering, Southeast University\\
$^2$Beijing Institute of Computer Technology and Application\\
\emails
\{gzli, zijiexu, ziyus1999, jiajliu, keji\}@seu.edu.cn
}

\begin{document}

\maketitle

\begin{abstract}
Dialogue relation extraction (DRE) aims to extract relations between two arguments within a dialogue, which is more challenging than standard RE due to the higher person pronoun frequency and lower information density in dialogues. However, existing DRE methods still suffer from two serious issues: (1) hard to capture long and sparse multi-turn information, and (2) struggle to extract golden relations based on partial dialogues, which motivates us to discover more effective methods that can alleviate the above issues. We notice that the rise of large language models (LLMs) has sparked considerable interest in evaluating their performance across diverse tasks. To this end, we initially investigate the capabilities of different LLMs in DRE, considering both proprietary models and open-source models. Interestingly, we discover that LLMs significantly alleviate two issues in existing DRE methods. Generally, we have following findings: (1) scaling up model size substantially boosts the overall DRE performance and achieves exceptional results, tackling the difficulty of capturing long and sparse multi-turn information; (2) LLMs encounter with much smaller performance drop from entire dialogue setting to partial dialogue setting compared to existing methods; (3) LLMs deliver competitive or superior performances under both full-shot and few-shot settings compared to current state-of-the-art; (4) LLMs show modest performances on inverse relations but much stronger improvements on general relations, and they can handle dialogues of various lengths especially for longer sequences.
\end{abstract}

\section{Introduction}
Relation extraction (RE) aims to identify relationships between arguments in text, including sentences~\cite{zhang2017tacred}, documents~\cite{yao-etal-2019-docred}, or dialogues~\cite{yu-etal-2020-dialogue}. However, as many relational facts span multiple sentences, sentence-level RE faces limitations in practice~\cite{yao-etal-2019-docred}. Therefore, cross-sentence RE, which aims to identify relations between two entities not in the same sentence, is vital for building knowledge bases from extensive corpora. As dialogues often exhibit cross-sentence relations, extracting relations from dialogues is necessary~\cite{yu-etal-2020-dialogue}.

\begin{figure}[t]
    \small
	\centering
	\includegraphics[width=\linewidth]{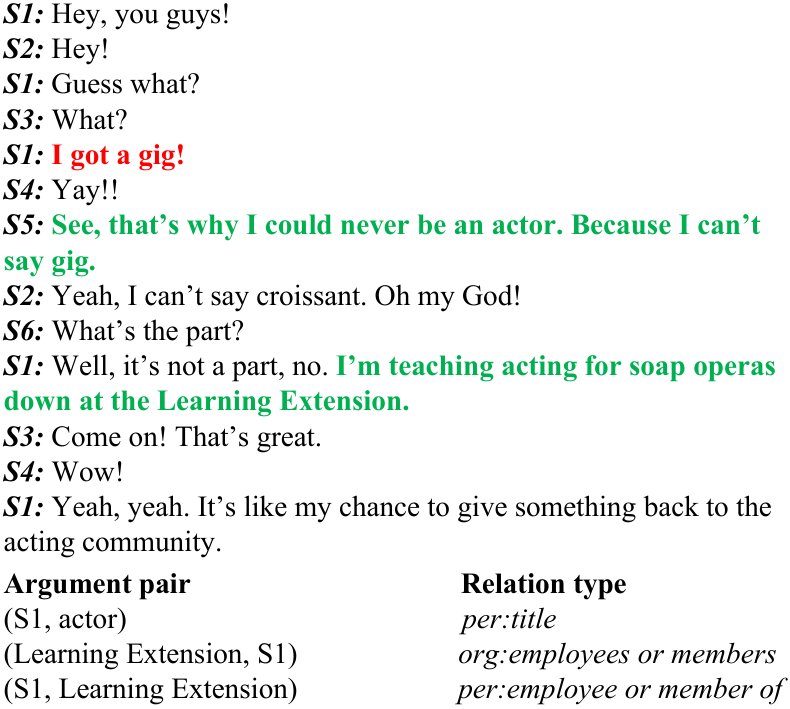}
	\caption{An example dialogue with its desired relations. S1-S6: anonymized speaker of each utterance.}
	\label{dialogue}
\end{figure}

To extract relations between two arguments that appear within a dialogue, dialogue relation extraction (DRE) models require reasoning over multiple sentences in a dialogue. To this end, a plethora of models have been proposed to address the challenges of DRE in recent studies including \textit{sequence-based} methods~\cite{yu-etal-2020-dialogue,son-etal-2022-grasp} and \textit{graph-based} methods~\cite{xue2021gdpnet,lee-choi-2021-graph,liu2023hierarchical}. Sequenced-based methods utilize encoder-only pre-trained language models such as BERT~\cite{devlin2019bert} to encode utterances in the dialogue. Graph-based methods usually construct various heterogeneous graphs over dialogues to capture the relational information. However, neither paradigm of methods can deliver satisfactory DRE results, because these encoder-based methods are sensitive to the higher person pronoun frequency and lower information density in dialogues~\cite{lee-choi-2021-graph}. As shown in Figure~\ref{dialogue}, the sentence highlighted with red is a prerequisite for correct understanding of relationships in the following sentences highlighted with green, but this sparse multi-turn information is difficult for models to capture. Moreover, current methods suffer grave performance declines when estimating how many turns it require to predict the relations between two arguments in partial dialogue instead of entire dialogue~\cite{yu-etal-2020-dialogue}. For example, DRE models are expected to deduce the relation \textit{per:title} between \textit{S1} and \textit{actor} with only first seven utterances in Figure~\ref{dialogue} while current methods typically need to rely on entire dialogue. This motivates us to search for more effective methods that can alleviate the above issues.

Besides the above sequence-based and graph-based methods, \textit{generation-based} methods are also occasionally used in RE, but only focus on sentence-level~\cite{cabot2021rebel,paolini2021structured} and document-level~\cite{giorgi2022sequence} RE. As the emergence of large language models (LLMs) such as GPT-3~\cite{brown2020language} significantly driving the progress of natural language processing (NLP), utilizing generation-based methods start to draw more attention~\cite{wadhwa2023revisiting,zhang2023aligning,ma2023large,li2023revisiting,wan2023gpt} in RE community. To the best of our knowledge, however, there is currently no work specifically applying generation-based methods to DRE. To this end, we aim to investigate the capabilities of generation-based methods for DRE in this work. Specifically, we consider the evaluation of both powerful proprietary LLMs such as ChatGPT~\cite{openai2022} and open-source LLMs such as LLaMA~\cite{touvron2023llama}. For proprietary LLMs, we design various prompt formats to extract relations from dialogues via few-shot prompting~\cite{brown2020language} and zero-shot prompting~\cite{kojima2022large}. For open-source models evaluation, we introduce \textbf{Landre} (\textbf{Lan}guage models driven \textbf{d}ialogue \textbf{r}elation \textbf{e}xtraction), a DRE framework driven by LLMs based on open-source foundation models, where it employs a simple prompt tuning method and a parameter efficient tuning technique LoRA~\cite{hu2021lora} for better RE task adaptation. We compare the performances of generation-based methods with previous sequence-based and graph-based methods, and conduct extensive experiments to provide valuable insights and guide further exploration.

Importantly, we empirically discover that generation-based methods significantly alleviate the issues of DRE mentioned above. On the one hand, for the difficulty of capturing long and sparse multi-turn information, we find that scaling up model size substantially boosts the overall DRE performance and achieves exceptional results. While the ChatGPT demonstrates promising performance with elaborated prompts, fine-tuned open-source models surpass prior state-of-the-art (SoTA) by a large margin, opening new frontiers for advancing the research of DRE. On the other hand, for the requirement of partial dialogue understanding, we surprisingly find that all generation-based methods encounter with much smaller performance drop from entire dialogue to partial dialogue setting, comparing to sequence-based and graph-based methods. In summary, our contributions are as follows:
\begin{itemize}
    \item We present the initial evaluation of LLMs regarding to DRE. We design elaborated prompts for ChatGPT and introduce an LLM-driven DRE framework Landre for open-source foundation models. Evaluation on DRE benchmarks across various experimental settings reveals our significant performance improvements over previous sequence-based and graph-based methods.
    \item We propose to utilize generation-based methods in DRE, which significantly alleviate two main issues of DRE. Generation-based methods are able to capture important relation semantics from sparse multi-turn information and achieve relatively consistent performances under both partial and entire dialogue settings.
    \item By analyzing the performance and complexity of different generation-based methods, we emphasize their promising performance and obvious shortcomings compared to previous methods, providing valuable insights to advance future DRE research.
\end{itemize}

\section{Dialogue Relation Extraction (DRE)}
Given a dialogue $D=s_1 : t_1, s_2 : t_2, ..., s_m : t_m$ and an argument pair $(a_1, a_2)$, where $s_i$ and $t_i$ denote the speaker ID and text of the $i^{th}$ turn, respectively, and $m$ is the total number of turns, we evaluate the performance of approaches in extracting relations between $a_1$ and $a_2$ that appear in $D$ in the following two settings. (1) \textbf{Standard Setting:} Similar to standard RE tasks, the dialogue $D$ is regarded as a document $d$, and we evaluate the extracted relations between $a_1$ and $a_2$ based on $d$. Specifically, we adopt F1, which is the harmonic mean of precision (P) and recall (R), for evaluation. (2) \textbf{Conversational Setting:} Instead of only considering the entire dialogue, we can regard the first $i \leq m$ turns of the dialogue as $d$.  Given $d$ containing the first $i$ turns in a dialogue, relation type $r$ associated with $a_1$ and $a_2$ is evaluable if $a_1$, $a_2$, and the trigger for $r$ have all been mentioned in $d$. The definition is based on the assumption that we can roughly estimate how many turns we require to predict the relations between two arguments based on the positions of the arguments and triggers. Accordingly, we adopt a new metric F1$_c$, the harmonic mean of conversational precision (P$_c$) and recall (R$_c$), as a supplement to the standard F1 metric~\cite{yu-etal-2020-dialogue}.

\section{Large Language Models for DRE}
We leverage various LLMs to address DRE task. We consider both proprietary LLMs such as ChatGPT~\cite{openai2022} and open-source LLMs such as LLaMA~\cite{touvron2023llama}.

\begin{figure*}[t]
    \small
	\centering
	\includegraphics[width=\linewidth]{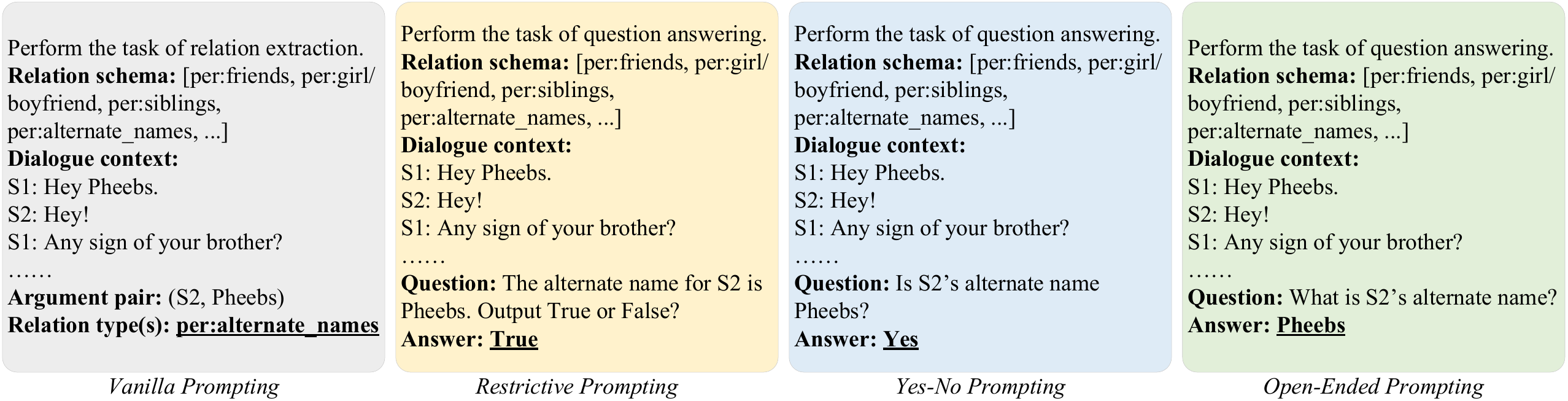}
	\caption{Formats of four prompting. The outputs of LLMs are highlighted with \underline{underline}.}
	\label{prompts}
\end{figure*}

\subsection{Leveraging ChatGPT for DRE}
We access ChatGPT using official API service~\footnote{https://platform.openai.com/} by prompting. As prompting is a brittle process wherein small modifications to the prompt can cause large variations in the model predictions, we explore various prompt templates to investigate the potential of ChatGPT for DRE in two distinct extraction manner: direct extraction and indirect extraction. 
\paragraph{Direct Extraction} Direct extraction makes ChatGPT directly output the relation labels between an given argument pair based on the input dialogue and candidate relations. This vanilla prompting method is simple, direct, and efficient.
\paragraph{Indirect Extraction} Indirect extraction decomposes the DRE task into two steps of asking then answering. Specifically, we sequentially ask ChatGPT if an argument pair expresses a certain relation among candidate relations based on the dialogue. We ground our analysis in three standard categories of prompts used
in prior work~\cite{brown2020language,sanh2021multitask,liu2024towards,arora2210ask,li2024unlocking,shang2024ontofact} including restrictive prompts, yes-no prompts and open-ended prompts. Restrictive prompts (e.g., ``S1 is an actor. Output True or False?'') construct questions that restrict the model output particular tokens, while yes-no prompts (e.g., ``Is S1 an actor?'') and open-ended prompts (e.g., ``What is the title of S1?'') construct free-form questions, as shown in Figure~\ref{prompts}. Typically, the indirect prompting method suffers from relatively high inference complexity as it needs to enumerate all possible triples to obtain questions and answers. Suppose we have $n$ samples to be extracted and $r$ candidate relations, the total complexity is $O (r \times n)$, which is time-consuming in practice. The purpose of experimenting with indirect extraction is to explore the DRE potential of LLMs as one step extraction regularly leads to suboptimal results~\cite{li2023revisiting}.

\subsection{Tuning Open-Source Models for DRE}
Different with the aforementioned usage of ChatGPT, we fine-tune the open-source LLMs for better task adaptation. Specifically, we introduce Landre, a DRE framework driven by LLMs based on smaller, open-source foundation models such as LLaMA~\cite{touvron2023llama}, where we investigate the performance of fine-tuning generative LLMs for DRE. We first outline the process of constructing input and output prompt formats. Then we utilize a parameter-efficient fine-tuning (PEFT) technique to train the foundation model with limited computational resources.

\subsection{Prompt Tuning}
\begin{figure}[t]
    \small
	\centering
	\includegraphics[width=\linewidth]{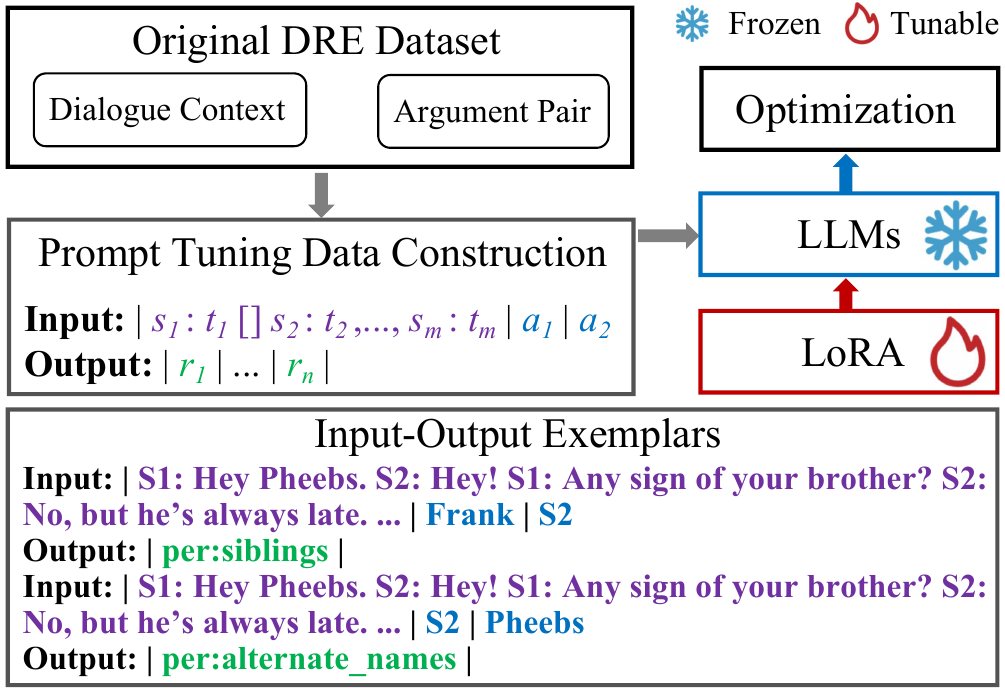}
	\caption{Paradigm of the Landre framework. In the first step, we construct the prompt tuning data from the original DRE dataset. Then we utilize the parameter-efficient fine-tuning technique LoRA to train the foundation model. The purple, blue and green texts in the above exemplars refer to dialogue context $D$, argument pair $(a_1, a_2)$ and relation set $R$, respectively.}
	\label{pt}
\end{figure}

As shown in Figure~\ref{pt}, for each sample $(D, a_1, a_2)$ in the training set, we construct the corresponding input prompt and let the LLM directly generate the corresponding relation label(s) indicating the relationship(s) between arguments $a_1$ and $a_2$ expressing by the dialogue $D$. Specifically, we concatenate the given dialogue and argument pair separated by the recognizable delimiter ``$|$'' as the input prompt, and we also separate the relation labels between the given argument pair by the ``$|$'' symbol as the output prediction. The detailed prompt tuning settings are introduced as follows.

The input prompt is composed of three parts: (1) the dialogue context $D=s_1 : t_1, s_2 : t_2, ..., s_m : t_m$, (2) the argument $a_1$ and (3) the argument $a_2$. And the output is a relation set $R=\{r_1,...,r_n\}$ where $n$ denotes the number of relations between argument pair $(a_1, a_2)$:
\begin{equation}
    | \, s_1 : t_1 [] s_2 : t_2 \, ... \, s_m : t_m \, | \, a_1 \, | \, a_2 \, \rightarrow | \, r_1 \, |\, ... \, | \, r_n \, |
\end{equation}
Note that in most cases, $n=1$.  For convenience, we concatenate all the speaker IDs and texts of multi-turn with the white space token $[]$, to indicate the start of each utterance. We empirically find that although this prompting strategy is frustratingly simple but shows its effectiveness.

\subsection{Parameter Efficient Tuning}
In this part, we describe how to fine-tune the foundation model using a parameter efficient approach. Landre takes the dialogue context and argument pair from the dataset as inputs and retrieves the corresponding relation label(s) as output:
\begin{equation}
    R = {\rm {Decoder}}(D, a_1, a_2)
\end{equation}
where Decoder indicates that the foundation model (e.g., LLaMA~\cite{touvron2023llama}) uses the Transformer-decoder framework. To enhance the efficiency of the fine-tuning process and reduce memory requirements, we utilize Low-Rank Adaptation (LoRA)~\cite{hu2021lora}, which freezes the pre-trained model weights and injects trainable rank decomposition matrices into each layer of the Transformer architecture, greatly reducing the number of trainable parameters for downstream tasks. Finally, the learning objective of the generation process is to minimize the negative loglikelihood of $R$ given the dialogue context $D$ and argument pair $(a_1, a_2)$:
\begin{equation}
    L = - \sum_{i}^{N} \sum_{j}^{|D_i|} \, \log p(R^j \, | \, D_i, a_1^j, a_2^j)
\end{equation}
where $N$ and $|D_i|$ denote the total number of dialogues in training set and argument pairs in dialogue $D_i$.

\section{Experiments}
\subsection{Experimental Settings}
\paragraph{Datasets and Metrics} We conduct experiments on the two versions (V1 and V2~\footnote{https://dataset.org/dialogre/}) of DialogRE dataset~\cite{yu-etal-2020-dialogue}, the first human-annotated DRE dataset, originating from the complete transcripts of the series \textit{Friends}. 
V1 is the original English version and V2 is the updated English version with a few annotation errors fixed. DialogRE has 36 relation types, 1,788 dialogues, and 8,119 argument pairs. The statistics of DialogRE are shown in Table~\ref{dialogre}.
We calculate both the F1 and F1$_c$~\cite{yu-etal-2020-dialogue} scores as the evaluation metrics. F1$_c$ is an evaluation metric to supplement the standard F1, computing by taking in the part of dialogue as input, instead of only considering the entire dialogue. We experiment Landre in both full-shot and few-shot settings. In the few-shot setting (V2), 8, 16, and 32-shot experiments are conducted by using three different randomly sampled data~\cite{son-etal-2022-grasp}. For the experiments on ChatGPT, to keep OpenAI API costs under control, we randomly select 100 dialogues from DialogRE valid and test set, respectively. For direct extraction, the evaluation is consistent with typical evaluation as the extracted relation labels are generated by ChatGPT. The indirect extraction could generate extra tokens that irrelavent with our targets and requires answer cleansing. Specifically, after the ChatGPT generation, we pick up only the part of the answer text that first satisfies answer format~\cite{kojima2022large}.

\begin{table}[t]
\small
\centering\setlength{\tabcolsep}{3mm}
\begin{tabular}{lccc}
\toprule
{DialogRE} & {Train} & {Dev} & {Test} \\
\midrule
{\# Conversations} & {1,073} & {358} & {357} \\
{Average dialogue length} & {229.5} & {224.1} & {214.2} \\
{\# Argument pairs} & {5,963} & {1,928} & {1,858} \\
{Average \# of turns} & {13.1} & {13.1} & {12.4} \\
{Average \# of speakers} & {3.3} & {3.2} & {3.3} \\
\bottomrule
\end{tabular}
\caption{Statistics of DialogRE.}
\label{dialogre}
\end{table}

\paragraph{Baseline Models} For a comprehensive performance evaluation, we compare the ChatGPT and Landre with both sequence-based and graph-based methods on the DialogRE dataset. Sequence-based methods usually utilize pre-trained language models to encode utterances in the dialogue, including BERT~\cite{devlin2019bert}, BERT$_s$~\cite{yu-etal-2020-dialogue}, RoBERTa$_s$~\cite{lee-choi-2021-graph} and GRASP~\cite{son-etal-2022-grasp}. BERT$_s$ is speaker-aware BERT. GRASP captures relational semantic clues of a given dialogue with the prompt marker strategy and the relational clue detection task. Graph-based methods include GDPNet~\cite{xue2021gdpnet}, TUCORE~\cite{lee-choi-2021-graph} and HiDialog~\cite{liu2023hierarchical}.  GDPNet develops a multiview latent graph to better capture the key features. TUCORE constructs a turn-level graph to capture the relational information. HiDialog leverages a heterogeneous graph module to polish the learned embeddings.

\paragraph{Experiment Details} For the ChatGPT, we consider both few-shot prompting~\cite{brown2020language} and zero-shot prompting~\cite{kojima2022large} for evaluation. For the zero-shot prompting, we utilize proposed four types of prompt formats. For the few-shot prompting, we only experiment with direct extraction (i.e. vanilla prompting), considering 1-shot, 3-shot and 5-shot settings because of the maximum length limitation of ChatGPT. And the few-shot demonstrations are randomly sampled from a uniform distribution over all training dialogue examples. Due to ICL is known to have high variance~\cite{zhao2021calibrate}, we compute the average performance of three times accessed results. We use the GPT-3.5 series model ``gpt-3.5-turbo'' for experiments.
For open-source LLMs, we experiment the Landre framework with generative models including GPT-2~\cite{radford2019language}, BART~\cite{lewis-etal-2020-bart}, T5~\cite{raffel2020exploring}, BLOOM~\cite{scao2022bloom} and LLaMA~\cite{touvron2023llama}. Specifically, we use the GPT-2-117M, BART-large-400M, BLOOM-560M, T5-large-770M and LLaMA-7B as the foundation models. We utilize LoRA~\cite{hu2021lora} to tune all these models for simplicity and efficiency. We set the rank $r$ of the LoRA parameters to 8 and the merging ratio $\alpha$ to 32. The model is optimized with AdamW~\cite{loshchilov2019decoupled} using learning rate 1e-4 with a linear warm up~\cite{goyal2017accurate} for the first 6\% steps followed by a linear decay to 0. We train Landre for 5 epochs with batch size 4, and all experiments are conducted on a single Geforce GTX 3090 GPU.

\begin{table*}[t]
\small
\centering\setlength{\tabcolsep}{1.5mm}
\begin{tabular}{c|c|c|cccc|cccc}
\toprule
\multirow{2}{*}{Type} & \multirow{2}{*}{Method} & {Based-model} & \multicolumn{2}{c}{\textbf{Dev (V1)}} & \multicolumn{2}{c}{\textbf{Test (V1)}} & \multicolumn{2}{c}{\textbf{Dev (V2)}} & \multicolumn{2}{c}{\textbf{Test (V2)}} \\
 &  & (\# Parameters) & {F1} & {F1{$_c$}} & {F1} & {F1{$_c$}} & {F1} & {F1{$_c$}} & {F1} & {F1{$_c$}} \\
\midrule
\multirow{4}{*}{Sequence-based} & {BERT~\cite{devlin2019bert}} & {BERT$_{\rm{base}}$ (110M)} & {60.6} & {55.4} & {58.5} & {53.2} & {59.4} & {54.7} & {57.9} & {53.1} \\
 & {BERT{$_s$}~\cite{yu-etal-2020-dialogue}} & {BERT$_{\rm{base}}$ (110M)} & {63.0} & {57.3} & {61.2} & {55.4} & {62.2} & {57.0} & {59.5} & {54.2} \\
 & {RoBERTa{$_s$}~\cite{lee-choi-2021-graph}} & {RoBERTa$_{\rm{large}}$ (355M)} & {-} & {-} & {-} & {-} & {72.6} & {65.1} & {71.3} & {63.7} \\
 & {GRASP~\cite{son-etal-2022-grasp}} & {RoBERTa$_{\rm{large}}$ (355M)} & {-} &{-} & {75.1} & {66.7} & {-} & {-} & {75.5} & {67.8} \\
\midrule
\multirow{3}{*}{Graph-based} & {GDPNet~\cite{xue2021gdpnet}} & {BERT$_{\rm{base}}$ (110M)} & {67.1} & {61.5} & {64.9} & {60.1} & {61.8} & {58.5} & {60.2} & {57.3} \\
 & {TUCORE~\cite{lee-choi-2021-graph}} & {RoBERTa$_{\rm{large}}$ (355M)} & {-} & {-} & {-} & {-} & {74.3} & {67.0} & {73.1} & {65.9} \\
 & {HiDialog~\cite{liu2023hierarchical}} & {RoBERTa$_{\rm{large}}$ (355M)} & \underline{76.9} & \underline{69.8} & \underline{77.7} & \underline{69.0} & \underline{76.4} & \underline{68.5} & {77.1} & {68.2} \\
\midrule
\multirow{3}{*}{Few-shot prompting} & {Vanilla Prompting (1-shot)} & {GPT-3.5 (200B)} & {59.8} & {57.8} & {61.0} & {58.6} & {59.2} & {57.5} & {60.6} & {58.4} \\
& {Vanilla Prompting (3-shot)} & {GPT-3.5 (200B)} & {61.9} & {58.8} & {62.3} & {59.6} & {61.7} & {58.2} & {62.0} & {58.9} \\
& {Vanilla Prompting (5-shot)} & {GPT-3.5 (200B)} & {62.2} & {58.9} & {62.9} & {60.4} & {61.2} & {58.3} & {62.6} & {59.7} \\
\midrule
\multirow{4}{*}{Zero-shot prompting} & {Vanilla Prompting (0-shot)} & {GPT-3.5 (200B)} & {57.8} & {55.4} & {59.5} & {57.0} & {57.4} & {55.7} & {58.9} & {56.3} \\
& {Restrictive Prompting (0-shot)} & {GPT-3.5 (200B)} & {74.6} & {72.4} & {74.2} & {70.3} & {75.3} & {72.0} & {74.9} & {70.8} \\
& {Yes-No Prompting (0-shot)} & {GPT-3.5 (200B)} & {75.2} & {72.4} & {73.5} & {71.7} & {75.7} & {72.6} & {74.2} & {71.2} \\
& {Open-Ended Prompting (0-shot)} & {GPT-3.5 (200B)} & {67.4} & {64.8} & {67.0} & {64.2} & {68.4} & {65.1} & {67.3} & {64.9} \\
\midrule
\multirow{4}{*}{Generation-based} & {GPT-2~\cite{radford2019language}} & {GPT-2 (117M)} & {55.9} & {54.0} & {56.8} & {54.3} & {54.6} & {52.5} & {55.3} & {53.7} \\
 & {BART~\cite{lewis-etal-2020-bart}} & {BART$_{\rm{large}}$ (400M)} & {61.3} & {59.8} & {61.8} & {60.2} & {60.7} & {59.5} & {61.0} & {59.8} \\
 & {BLOOM~\cite{scao2022bloom}} & {BLOOM (560M)} & {69.1} & {67.8} & {69.8} & {68.2} & {68.3} & {67.0} & {69.1} & {67.4} \\
 & {T5~\cite{raffel2020exploring}} & {T5$_{\rm{large}}$ (770M)} & {72.4} & {70.4} & {72.8} & {70.5} & {70.9} & {69.0} & {71.6} & {69.8} \\
 & {LLaMA~\cite{touvron2023llama}} & {LLaMA (7B)} & \textbf{78.2} & \textbf{77.1} & \textbf{79.4} & \textbf{78.5} & \textbf{77.8} & \textbf{76.6} & \textbf{79.0} & \textbf{77.9} \\
\bottomrule
\end{tabular}
\caption{Performance of ChatGPT and Landre on DialogRE in the full-shot setting, averaged over three runs. Best results are \textbf{bold} and our re-implemented results are marked with \underline{underline}.}
\label{landre}
\end{table*}

\subsection{Main Results of ChatGPT}
The main results on DialogRE using ChatGPT as relation extractors are shown in Table~\ref{landre}. We have the following findings:

\paragraph{Indirect extraction tends to show much better performance than direct extraction for ChatGPT.} Obviously, the vanilla prompting, which makes ChatGPT directly output the relation labels between an given argument pair, achieves the worst F1 scores in both few and zero-shot prompting manners compared to other three zero-shot prompting formats. Decomposing extraction process into multiple steps can improve the zero-shot performance by more than 15\% in absolute F1 scores. Different with existing findings on prompting formats~\cite{arora2210ask}, open-ended prompting tends to perform worse than restrictive and yes-no prompting in DRE. In addtion, for few-shot prompting results, the improvements regarding to zero-shot prompting are small, where 5-shot results seem to have limited advantages compared to 3-shot.

\paragraph{ChatGPT surpasses several strong sequence-based and graph-based methods.} Even with the zero-shot prompting, ChatGPT with elaborated prompting formats delivers better results than fully supervised RoBERTa$_s$ (+2.9\%$\sim$7.5\%) and TUCORE-GCN (+1.1\%$\sim$5.6\%). Although ChatGPT is slightly worse than previous SoTA HiDialog (-0.7\%$\sim$3.5\%) in standard setting, we should note that our limited budget restricted our study to a small set of prompt styles. It is possible that having a larger prompt design search space could narrow the performance gap between ChatGPT and HiDialog.

\paragraph{Performance gap between standard and conversational settings for ChatGPT is less than previous methods.} Previous methods suffer grave declines when encountering conversational setting compared to standard setting (e.g., HiDialog drops around 7.1\%$\sim$8.9\%). Interestingly, ChatGPT significantly narrows this performance gap (less than 5\%), indicating ChatGPT can effectively identify the relations between an argument pair while tolerating the variant length of input dialogues. In contrast, encoder-only methods are sensitive to the length of input dialogues and hard to recover good representations for classification, because conversational setting is actually make the data distribution change in testing stage. 

\paragraph{ChatGPT still has limitations.} In summary, ChatGPT exhibits comparable performance in DRE compared to the previous SoTA methods, which highlights the ability of current LLMs to capture and comprehend complex linguistic patterns and dependencies within multi-turn dialogues. However, ChatGPT still faces two notable limitations. First, achieving good DRE performance requires elaborated prompting methods, resulting in higher computation cost and time consumption while ChatGPT is subject to request limitations. Besides, responses from ChatGPT frequently contain substantial explanatory content, where it may not align perfectly with expected answer format and difficult to cleanse and evaluate. Second, ChatGPT is not open-source, restricting its modification and customization, raising concerns about privacy protection and feasibility of local deployment. Therefore, we propose to use open-source LLMs in DRE, which alleviates two issues in DRE but also avoids the limitations of ChatGPT.

\subsection{Main Results of Open-Source Models}
\paragraph{Full-shot Setting} We show the performance of Landre with different foundation models on the DialogRE dataset in Table~\ref{landre} compared with other baselines. It is shown that Landre achieves competitive or superior performance compared to all of the previous methods, including the current SoTA methods GRASP and HiDialog. On the one hand, scaling up the foundation models in Landre brings significant performance improvements. From 117M to 7B, generation-based methods demonstrate over 20\% absolute F1 gains in both standard and conversational settings. Note that the significant improvements are also observed in encoder-based methods (from 110M to 355M), but as the current encoder scale is much smaller than the decoder, the generation-based methods become more promising in future DRE. On the other hand, Landre shows its efficiency in conversational settings by narrowing the performance gap between F1 and F1$_c$ (e.g., Landre with LLaMA only drops around 0.9\%$\sim$1.2\% compared to 7.1\%$\sim$8.9\% in HiDialog), thereby indicating that generation-based methods effectively overcome the low information density of dialogues. Similar with ChatGPT, generation-based methods tend to perform robust encountering various length of input dialogues, and they are able to extract the golden relation based on partial dialogues with enough arguments and relation information.
These results imply that guiding the model to pay attention to relation labels with a prompt tuning approach can be more effective than adding additional features and layers in encoder-based methods.

\begin{table}[t]
\small
\centering\setlength{\tabcolsep}{1.5mm}
\begin{tabular}{ccccc}
\toprule
\multirow{2}{*}{Method} & {Based-model} & \multicolumn{3}{c}{\textbf{Shot}} \\
& {(\# Parameters)} & {K=8} & {K=16} & {K=32} \\
\midrule
{RoBERTa} & {RoBERTa$_{\rm{large}}$ (355M)} & {29.8} &{40.8} & {49.7} \\
{TUCORE} & {RoBERTa$_{\rm{large}}$ (355M)} & {24.6} & {40.0} & {53.8} \\
{GRASP$_{\rm{base}}$} & {RoBERTa$_{\rm{base}}$ (125M)} & \textbf{45.4} & {52.0} & {56.0} \\
{GRASP$_{\rm{large}}$} & {RoBERTa$_{\rm{large}}$ (355M)} & {36.0} & {55.3} & {62.6} \\
\midrule
{GPT-2} & {GPT-2 (117M)} & {38.5} & {44.2} & {54.0} \\
{BART} & {BART$_{\rm{large}}$ (400M)} & {36.4} & \textbf{56.9} & {58.8} \\
{BLOOM} & {BLOOM (560M)} & {30.3} & {53.6} & {60.2} \\
{T5} & {T5$_{\rm{large}}$ (770M)} & {29.6} & {52.0} & {61.9} \\
{LLaMA} & {LLaMA (7B)} & {26.7} & {48.2} & \textbf{65.4} \\
\bottomrule
\end{tabular}
\caption{Low-resource DRE performance of F1 scores. We use K = 8, 16, 32 (\# of examples per relation) for few-shot experiments.}
\label{fewshot}
\end{table}

\paragraph{Few-shot Setting} As presented in Table~\ref{fewshot}, Landre exhibits robust performance in few-shot settings. For example, Landre with GPT-2 surpasses TUCORE regardless of the number of shots. Except for the 8-shot setting, Landre presents new SoTA performance compared to previous methods. However, GRASP with smaller encoder still delivers better 8-shot performance compared to others. Moreover, although Landre with LLaMA is much better than other baselines in the 32-shot setting, the limited performance of LLaMA in the 8-shot and 16-shot setting can be attributed to an insufficient number of examples for tuning large-scale parameters. 

\section{Analysis}
In this section, we analyze various aspects of Landre to provide a deeper understanding of generation-based methods performing in DRE task. Specifically, we analyze the results of GPT-2 and LLaMA on the DialogRE (V2) dev and test set.

\subsection{Error Analysis}
We calculate the error rate of each relation to investigate the fine-grained performance of Landre with GPT-2 and LLaMA. To mitigate the deviation caused by imbalanced relation types, we analyze the 10 most frequently appearing relations whose corresponding sample numbers are above 100. The overall results are summarized in Table~\ref{error}, where scaling up model size consistently reduces the error rates of all relations. For most interpersonal relations that frequently appear such as per:girl/boyfriend, per:spouse and per:children, LLaMA delivers prominent performance improvements compared to GPT-2, showing increasing the model size can better capture intra-turn and inter-turn information, distinguishing the relationship semantic differences in the dialogues.

For specific relations such as per:alternate\_names and per:title, GPT-2 and LLaMA achieve similar precision with low error rates. First, it is not surprising that the relation per:alternate\_names is easy for models because it is the most frequently appearing relation and typically can be identified within a single-turn information (e.g., S1 calls S2's name then S2 responds). For relation per:title, interestingly, two models have the same number of wrong predictions. We compare the error cases of two models and find 5 out of 6  samples overlapping. The 5 samples are mistakenly annotated as per:title between PER and NAME in the first place while the subject and object types of per:title should be PER and STRING~\cite{yu-etal-2020-dialogue}. Because most of argument pair types  belong to PER and NAME, per:title is relatively easy to distinguish. In other words, generative methods can achieve near perfect performance on relation per:title because of the information from argument mentions, most of which is type information.

\begin{table}[t]
\small
\centering\setlength{\tabcolsep}{1mm}
\begin{tabular}{c|c|cc|cc}
\toprule
\textbf{Relation label} & \textbf{Triple} & \textbf{GPT-2} & \textbf{ER.} & \textbf{LLaMA} & \textbf{ER.} \\
\midrule
{per:alternate\_names} & {819} & {59} & {7.2} & {35} & {4.3} \\
{per:girl/boyfriend} & {306} & {160} & {52.3} & {83} & {27.1} \\
{per:positive\_impression} & {287} & {208} & {72.5} & {81} & {28.2} \\
{per:friends} & {278} & {63} & {22.7} & {57} & {20.5} \\
{per:title} & {164} & {6} & {3.7} & {6} & {3.7} \\
{per:spouse} & {126} & {81} & {64.3} & {31} & {24.6} \\
{per:siblings} & {122} & {49} & {40.2} & {26} & {21.3} \\
{per:children} & {103} & {88} & {85.4} & {25} & {24.3} \\
{per:parents} & {103} & {60} & {58.3} & {37} & {35.9} \\
{per:negative\_impression} & {102} & {102} & {100.0} & {51} & {50.0} \\
\bottomrule
\end{tabular}
\caption{Error rates of Top-10 relations on the DialogRE dev and test set. \textbf{Triple} denotes the total number of triples of each relation in the dev and test set. \textbf{GPT-2} and \textbf{LLaMA} denote the number of wrong predictions and \textbf{ER.} denotes the corresponding error rate.}
\label{error}
\end{table}

Another interesting finding is that GPT-2 delivers 100\% error rate on relation per:negative\_impression. We thus examine the proportion of its predicted results with respect to 102 samples annotated with per:negative\_impression. We discover that 66.6\% of samples are predicted with relations per:girl/boyfriend, per:positive\_impression and per:friend for GPT-2, where these relations may share same argument pairs with per:negative\_impression and tend to confuse the models because friends and girl/boyfriends often express impressions of each other. In addition, some specific relations such as per:acquaintance (not shown in Table~\ref{error}) are challenging to Landre, where the error rates of per:acquaintance are 100\% and 97.8\% for GPT-2 and LLaMA, respectively. The main reasons include difficulty in defining such relation (i.e., it is difficult to distinguish between acquaintances and friends) and low trigger ratio (around 22\%)~\cite{yu-etal-2020-dialogue}.

\subsection{Qualitative Analysis}

\begin{table}[t]
\small
\centering\setlength{\tabcolsep}{3mm}
\begin{tabular}{cccc}
\toprule
\textbf{Method} & \textbf{I} & \textbf{II} & \textbf{III} \\
\midrule
{BERT~\cite{devlin2019bert}} & {42.5} & {60.7} & {65.6} \\
{GDPNet~\cite{xue2021gdpnet}} & {47.4} & {59.8} & {68.1} \\
{RoBERTa$_{\rm{s}}$~\cite{lee-choi-2021-graph}} & {57.4} & {69.3} & {79.6} \\
{TUCORE~\cite{lee-choi-2021-graph}} & {62.3} & \textbf{71.3} & {79.9} \\
{HiDialog~\cite{liu2023hierarchical}} & \textbf{76.6} & {70.5} & {80.9} \\
\midrule
{GPT-2~\cite{radford2019language}} & {39.0} & {43.8} & {72.2} \\
{LLaMA~\cite{touvron2023llama}} & {75.3} & {69.3} & \textbf{88.2} \\
\bottomrule
\end{tabular}
\caption{All methods performance on DialogRE. We break down the performance into three groups (I) asymmetric inverse relations, (II) symmetric inverse relations, and (III) others.}
\label{inverse}
\end{table}

\begin{figure}[ht]
    \small
	\centering
	\includegraphics[width=\linewidth]{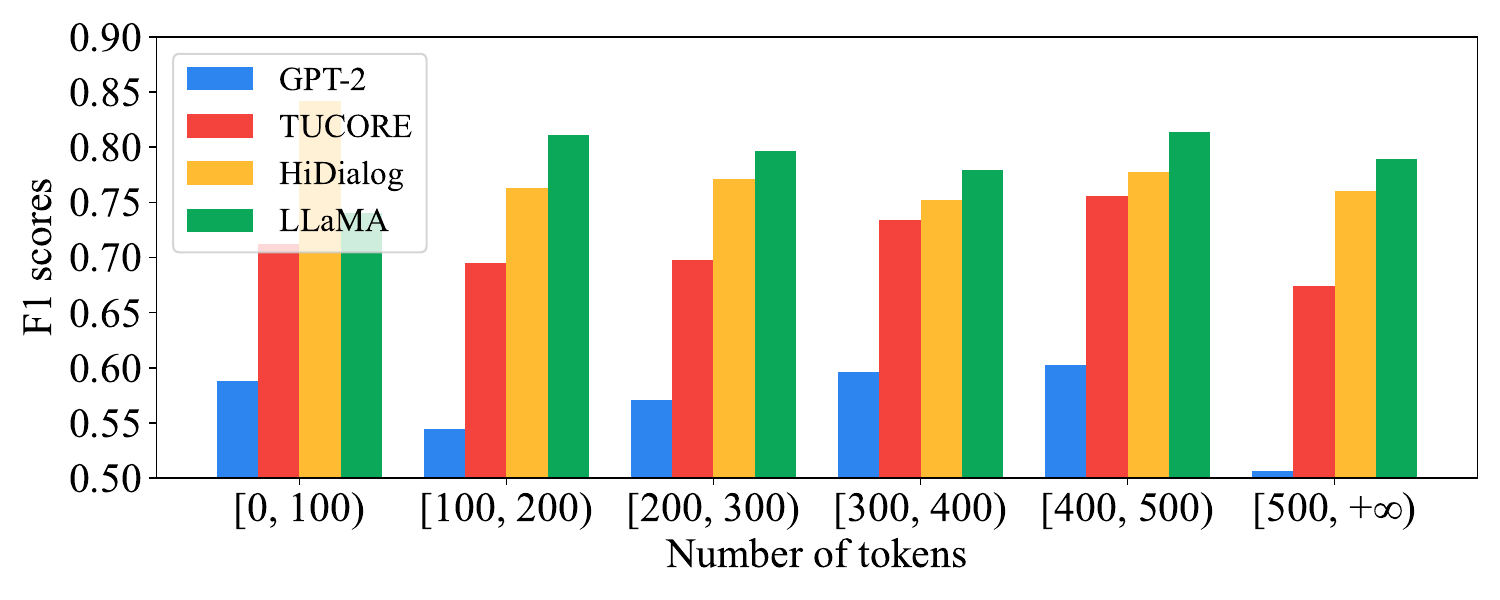}
	\caption{Analysis of robustness of Landre tackling increasing utterance length compared to baseline TUCORE and HiDialog.}
	\label{tokens}
\end{figure}

\paragraph{Inverse relations.} We group the test set of DialogRE according to the relation types into three subsets: (I) asymmetric, when a relation type differs from its inversion (e.g. per:children and per:parents); (II) symmetric, when a relation type is the same as its inversion (e.g. per:friends); (III) other, when a relation type does not have inversion (e.g. per:title). We compare the performance of Landre with
baselines and report the results in Table~\ref{inverse}. We observe that GPT-2 struggles to capture inverse relations compared to BERT. Moreover, Landre with LLaMA shows comparable performance on both asymmetrical and symmetrical relations compared to TUCORE and HiDialog, while it delivers exceptional results on other relation types. And this is also why GPT-2 and LLaMA have low error rates on specific relations such as per:alternate\_names which is the most frequently appearing relation and contributes the majority of overall performance.

\paragraph{Utterance lengths.} Compared to encoder-based methods, generation-based methods enable to handle dialogues of various lengths. We further divide the samples in the test set into six groups according to their lengths and report the F1 score for each group achieved by Landre and previous SoTA. As shown in Figure~\ref{tokens}, LLaMA outperforms HiDialog in all groups except the group with less than 100 tokens, maintaining decent performance for long sequences.

\begin{table}[b]
\small
\centering\setlength{\tabcolsep}{3mm}
\begin{tabular}{ccc}
\toprule
\textbf{Method} & \textbf{MELD} & \textbf{EmoryNLP} \\
\midrule
{RoBERTa~\cite{liu2019roberta}} & {62.0} & {37.3} \\
{TUCORE~\cite{lee-choi-2021-graph}} & {65.4} & {39.2} \\
{GRASP~\cite{son-etal-2022-grasp}} &{65.6} & {40.0} \\
{HiDialog~\cite{liu2023hierarchical}} &{65.7} & {38.1} \\
\midrule
{LLaMA~\cite{touvron2023llama}} & \textbf{66.2} & \textbf{40.7} \\
\bottomrule
\end{tabular}
\caption{Experimental results of LLaMA on MELD and EmoryNLP.}
\label{new}
\end{table}

\subsection{Applicability Analysis}
To evaluate the robustness and generalization of generation-based methods, we evaluate Landre with LLaMA on MELD~\cite{poria-etal-2019-meld} and EmoryNLP~\cite{zahiri2018emotion} datasets, which are designed for emotion recognition in conversations (ERC). Following previous works~\cite{lee-choi-2021-graph,son-etal-2022-grasp}, the ERC task is converted into DRE and the weighted-F1 is calculated to evaluate the MELD and EmoryNLP datasets. As presented in Table~\ref{new}, the performance of Landre surpasses that of GRASP and HiDialog in both MELD and EmoryNLP, demonstrating significant promise in dialogue understanding.

\section{Related Work}
\paragraph{Large Language Models.} Besides the “pre-train and fine-tune” paradigm~\cite{devlin2019bert}, pre-trained LLMs possess characteristics that are advantageous for few-shot~\cite{brown2020language} and zero-shot~\cite{kojima2022large} learning, whereby appropriate prompts are used to effectively guide the model towards generating desired task outputs, thus beginning an era of “pre-train and prompt”. However, prompting without updating parameters generally brings suboptimal results and the efficient utilization of LLMs remains a significant challenge. Recent advancements in parameter-efficient fine-tuning (PEFT) techniques have effectively alleviated this problem, such as LoRA~\cite{hu2021lora} and Prefix Tuning~\cite{liu-etal-2022-p}. In this study, we initially investigate the capabilities of fine-tuned open source (e.g. LLaMA~\cite{touvron2023llama}) and non fine-tuned closed source (e.g. ChatGPT~\cite{openai2022}) LLMs for DRE.

\paragraph{Dialogue Relation Extraction.} Most previous works on RE focused on sentence-level RE~\cite{li2022fastre,wang2023pascore,wang2023fmlre}, but recently cross-sentence RE has been studied more because many relational facts are expressed in multiple sentences in practice~\cite{yao-etal-2019-docred}.
Dialogue relation extraction (DRE) requires reasoning over multiple sentences in a dialogue to predict the relations between two arguments~\cite{yu-etal-2020-dialogue}. Current popular DRE methods can be divided into sequence-based methods~\cite{yu-etal-2020-dialogue,son-etal-2022-grasp} and graph-based methods~\cite{xue2021gdpnet,lee-choi-2021-graph,liu2023hierarchical}. The former utilize encoder-only language models such as BERT~\cite{devlin2019bert} and RoBERTa~\cite{liu2019roberta} to encode utterances in the dialogue. The latter construct various heterogeneous graphs over dialogues to capture the relational information. And both two kind of methods use the softmax classifiers for relation classification. In this study, we aim to investigate the capabilities of generative methods for DRE. Due to the rapid development of decoder-based LLMs, we anticipate that Landre could serve as a compelling baseline or plug-in module for future work in the community.

\section{Conclusion}
In this study, we conduct an initial examination of generative LLMs in DRE, showcasing its excellence over previous sequence-based and graph-based methods. To solve the limitations of ChatGPT, we present Landre, an LLM-driven DRE framework based on smaller, open-source foundation models. By simply prompting tuning with LoRA, Landre delivers better performance than ChatGPT and achieves new state-of-the-art performance on DRE benchmarks. Comprehensive evaluations across different experimental settings showcase the unique characteristics of generation-based methods, opening new frontiers for advancing the research of DRE.

\section*{Acknowledgments}
We thank the reviewers for their insightful comments. This work was supported by National Science Foundation of China (Grant Nos.62376057) and the Start-up Research Fund of Southeast University (RF1028623234). All opinions are of the authors and do not reflect the view of sponsors.

\bibliographystyle{named}
\bibliography{ijcai24}

\end{document}